\documentclass[11pt,a4paper]{article}
\usepackage{times,latexsym}
\usepackage{url}
\usepackage[T1]{fontenc}

\usepackage[acceptedWithA]{tacl2021v1}

\usepackage{xspace,mfirstuc,tabulary}

\newif\iftaclinstructions
\taclinstructionsfalse 
\iftaclinstructions
\renewcommand{\confidential}{}
\renewcommand{\anonsubtext}{(No author information supplied here, for consistency with
TACL-submission anonymization requirements)}
\newcommand{\instr}
\fi

\iftaclpubformat 

\else

\fi

\usepackage{paralist}
\usepackage{graphicx}
\usepackage{amssymb}
\usepackage{amsmath,cases}
\usepackage{natbib}
\usepackage{multirow}
\usepackage{booktabs}
\usepackage{subfigure}
\usepackage{epstopdf}
\usepackage{comment}
\usepackage{url}
\usepackage{xcolor}
\usepackage{bbm}
\usepackage{color}
\usepackage{soul}

\usepackage{array}
\newcolumntype{L}[1]{>{\raggedright\let\newline\\\arraybackslash\hspace{0pt}}m{#1}}
\newcolumntype{C}[1]{>{\centering\let\newline\\\arraybackslash\hspace{0pt}}m{#1}}
\newcolumntype{R}[1]{>{\raggedleft\let\newline\\\arraybackslash\hspace{0pt}}m{#1}}

\title{Holistic Exploration on Universal Decompositional Semantic Parsing: Architecture, Data Augmentation, and LLM Paradigm}

\author{
  Hexuan Deng\textsuperscript{\rm 1},
  Xin Zhang\textsuperscript{\rm 1},
  Meishan Zhang\textsuperscript{\rm 1},
  Xuebo Liu\textsuperscript{\rm 1},
  Min Zhang\textsuperscript{\rm 1}
  \\
  \textsuperscript{\rm 1} Institute of Computing and Intelligence, Harbin Institute of Technology, Shenzhen, China
  \\
  \texttt{\{22s051030,zhangxin2023\}@stu.hit.edu.cn}
  \\
  \texttt{\{zhangmeishan,liuxuebo,zhangmin2021\}@hit.edu.cn}
}

\date{}

\begin{document}
\maketitle
\begin{abstract}
In this paper, we conduct a holistic exploration of the Universal Decompositional Semantic (UDS) Parsing. We first introduce a cascade model for UDS parsing that decomposes the complex parsing task into semantically appropriate subtasks. Our approach outperforms the prior models, while significantly reducing inference time. We also incorporate syntactic information and further optimized the architecture. Besides, different ways for data augmentation are explored, which further improve the UDS Parsing. Lastly, we conduct experiments to investigate the efficacy of ChatGPT in handling the UDS task, revealing that it excels in attribute parsing but struggles in relation parsing, and using ChatGPT for data augmentation yields suboptimal results. Our code is available at~\url{https://github.com/hexuandeng/HExp4UDS}.
\end{abstract}

\section{Introduction}

A long-standing objective in the fields of natural language understanding and computational semantics is to create a structured graph of linguistic meaning. Various efforts have been made to encode semantic relations and attributes into a semantic graph---e.g., Abstract Meaning Representation (AMR; \citealp{AbstractMeaningRepresentation_2013}), Universal Conceptual Cognitive Annotation (UCCA; \citealp{UniversalConceptualCognitive_2013}), and Semantic Dependency Parsing formalisms (SDP; \citealp{SemEval2014Task_2014a, ComparabilityLinguisticGraph_2016a}). Recently, Universal Decompositional Semantics (UDS; \citealp{UniversalDecompositionalSemantics_2020}) introduce an alternative approach, as shown in Figure~\ref{fig:dataset}. It constructs semantic relations from syntactic annotations \citep{EvaluationPredPattOpen_2017a}, and annotates semantic attributes following the decompositional semantics \cite{SemanticProtoRoles_2015a}, which takes the form of many simple questions about words or phrases, thus significantly lowering the annotation cost.

Previous parsing models for UDS dataset are mainly under the Seq2Seq transduction framework \citep{UniversalDecompositionalSemantic_2020}, which suffer from poor parallelism and long inference time that increase with the sentence length. In this paper, we propose a cascade architecture that decomposes the complex parsing task into multiple subtasks in a semantically appropriate manner. Within each subtask, our model predicts all corresponding sentence elements simultaneously, enhancing parallelism and substantially reducing inference time. Experimental results demonstrate that our approach outperforms previous models while maintaining high efficacy during inference.

To further improve our proposed model, we introduce enhancements from two perspectives. Firstly, we try to incorporate syntactic information, which has been proven beneficial to many downstream tasks \citep{AdaptiveKnowledgeSharing_2018b, SyntaxAwareOpinionRole_2020a, JointUniversalSyntactic_2021, Deng_Ding_Liu_Zhang_Tao_Zhang_2023}. We use multi-task training \cite{MultitaskLearning_1997a} as the default setting and propose several approaches for better utilizing syntactic information. Secondly, while \citet{UniversalDecompositionalSemantic_2020} have tried to utilize the external tool PredPatt \citep{EvaluationPredPattOpen_2017a}, which contains the relationship between syntax and semantic information, they do not achieve any improvements. In contrast, we propose a data augmentation method that effectively exploits the capabilities of PredPatt, leading to significant performance gains in relation parsing. Moreover, we have explored various approaches for these enhancements, providing guidance for the design of similar systems.

Large language models (LLMs), such as ChatGPT and GPT-4 \citep{SparksArtificialGeneral_2023}, have attracted considerable attention due to their impressive ability. These models engage in conversational interactions with users, accepting natural language prompts and producing textual responses. Their applications cover a broad spectrum, including machine translation \citep{ChatGPTGoodTranslator_2023}, grammar error correction \cite{ChatGPTGrammarlyEvaluating_2023}, information extraction \cite{EvaluatingChatGPTInformation_2023}, among others. We investigate the performance of LLMs on the UDS task. Our experiments involve either directly applying the LLMs for parsing or using LLMs to generate data to enhance downstream models. Results demonstrate that LLMs excel in attribute parsing but struggle in relation parsing, which appears to be too complex for LLMs.

\section{Background and Related Work}
\label{sec:related}

\paragraph{UDS Datasets} \citet{GoldStandardDependency_2014a} create a standard set of Stanford dependency annotations for the English Web Treebank (EWT, \citealp{GoldStandardDependency_2014a}) corpus. Subsequently, \citet{UniversalDecompositionalSemantics_2016} proposed a framework aimed at constructing and deploying cross-linguistically robust semantic annotation protocols and proposed annotations on top of the EWT corpus using PredPatt \citep{UniversalDecompositionalSemantics_2016, EvaluationPredPattOpen_2017a}. Several works have then been proposed to provide semantic annotations within this framework, including annotations for semantic roles \citep{SemanticProtoRoles_2015a}, entity types \citep{UniversalDecompositionalSemantics_2016}, event factuality \citep{NeuralModelsFactuality_2018}, linguistic expressions of generalizations about entities and events \citep{DecomposingGeneralizationModels_2019a}, and temporal properties of relations between events \citep{FineGrainedTemporalRelation_2019a}. All of these efforts culminated in \citet{UniversalDecompositionalSemantics_2020}, which presents the first unified decompositional semantics-aligned dataset, namely, Universal Decompositional Semantics (UDS).

\paragraph{UDS Parser} UDS parsing has been conducted using transition-based parser \citep{FastAccurateDependency_2014a}, deep biaffine attention parser \citep{DeepBiaffineAttention_2017a}, and sequence-to-graph transductive parser \citep{UniversalDecompositionalSemantic_2020}. The latter significantly outperforms the others by employing an efficient Seq2Seq transduction framework \citep{SequenceSequenceLearning_2014, NeuralMachineTranslation_2015}. This approach is initially used in AMR parsing \citep{AMRParsingSequencetoGraph_2019} and later extended to cover other semantic frameworks, such as UCCA and SDP, by \citet{BroadCoverageSemanticParsing_2019} in a unified transduction framework, which predicted nodes and corresponding edges simultaneously in a Seq2Seq manner. For UDS, an attribute module is added by \citet{UniversalDecompositionalSemantic_2020}. Syntactic information is incorporated into the model by \citet{JointUniversalSyntactic_2021}, yielding further improvements. Despite these attempts, cascade models with better parallelism and shorter inference time have not yet been explored.

\paragraph{Incorporating Syntactic Information} Syntactic information has been shown to improve the performance of downstream tasks. Multi-task learning is widely used to incorporate syntactic information. \citet{MultitaskParsingSemantic_2018a} improve the performance of semantic parsing by using multi-task learning, with syntactic and other semantic parsing tasks serving as auxiliary tasks. \citet{AdaptiveKnowledgeSharing_2018b} use syntactic and semantic information to improve the efficacy of two low-resource translation tasks. \citet{JointUniversalSyntactic_2021} employ a single model to parse syntactic and semantic information simultaneously to improve semantic parsing. Graph convolutional networks (GCN, \citealp{SemiSupervisedClassificationGraph_2017a}) have also been widely used. \citet{EncodingSentencesGraph_2017a} use GCNs to incorporate syntactic information in neural models and construct a syntax-aware semantic role labeling model. \citet{GraphConvolutionPruned_2018a} propose an extension of GCNs to help relation extraction models capture long-range relations between words. \citet{SyntaxAwareOpinionRole_2020a} present a syntax-aware approach based on dependency GCNs to improve opinion role labeling tasks.

\section{Preliminaries}
\label{sec:prelim}

The UDS dataset comprises three layers of annotations: syntactic annotations, semantic relation annotations, and decompositional semantic attribute annotations at the edge and node levels.

\begin{figure}[t]
    \centering
    \includegraphics[width=0.48\textwidth]{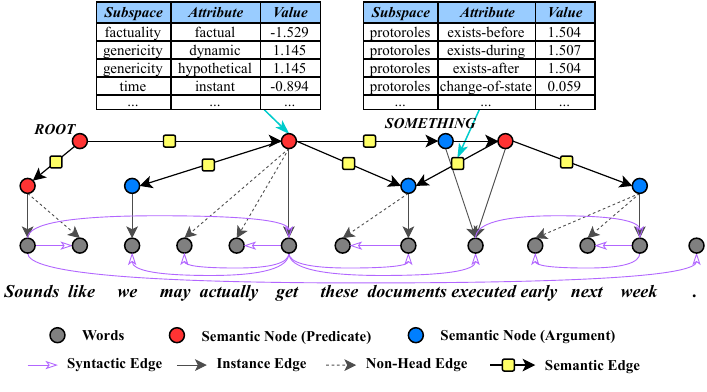}
    \vspace{-22pt}
    \caption{An example of UDS datasets with syntactic tree and semantic graph. Syntactic tree corresponds to the gray nodes and purple edges, semantic relations correspond to the red and blue nodes as well as the yellow edges, and semantic attributes are in the tables.}
    \label{fig:dataset}
\end{figure}

\paragraph{Syntactic Annotation} is derived from the EWT dataset, which provides consistent annotation of grammar, including part-of-speech (\emph{POS}) tag, morphological features, and syntactic dependencies, for human languages. These annotations are used to construct the \emph{syntactic tree}, where each word is tied to a node. As shown in Figure~\ref{fig:dataset}, the headword of "we" is "get", and the headword of the root "Sounds" is defined as itself.


\paragraph{Semantic Relations} consist of predicates, arguments, and edges between them, which forms the \emph{semantic relations}. It is generated by Predpatt tool \citep{EvaluationPredPattOpen_2017a} automatically, using the POS tag and the syntactic tree as input. Each semantic node explicitly corresponds to one word in the sentence called the center word, demonstrated by the instance edge. Additionally, each semantic node is also tied with several non-repetitive words with the non-head edge, which forms a multi-word span. As shown in Figure~\ref{fig:dataset}, the leftmost predict node has a span "Sounds like" with the center word "Sounds". Note that two semantic nodes may correspond to the same word in the case of clausal embedding. Then, an extra argument node "SOMETHING" is introduced as the the root of clause, e.g., "executed" is corresponding to an extra argument node.


\paragraph{Semantic Attributes} consist of crowdsourced decompositional annotations tied to the semantic relations, detailed in \S\ref{sec:related}. These annotations can be further categorized into node-level and edge-level attributes, corresponding to the table on the left and right in Figure~\ref{fig:dataset}, respectively. For each node or edge, all attributes have a value in range $[-3, 3]$. Besides, each attribute also has a confidence in range $[0, 1]$, which shows how likely it is to have the property. Following \citet{UniversalDecompositionalSemantic_2020}, we discretized it into $\{0, 1\}$ by setting every non-zero confidence to one.

\section{Methodology}

In this section, we introduce our cascade model, methods to improve its performance, and the exploration of LLMs.

\begin{figure*}[t]
	\centering
	\includegraphics[width=0.94\textwidth]{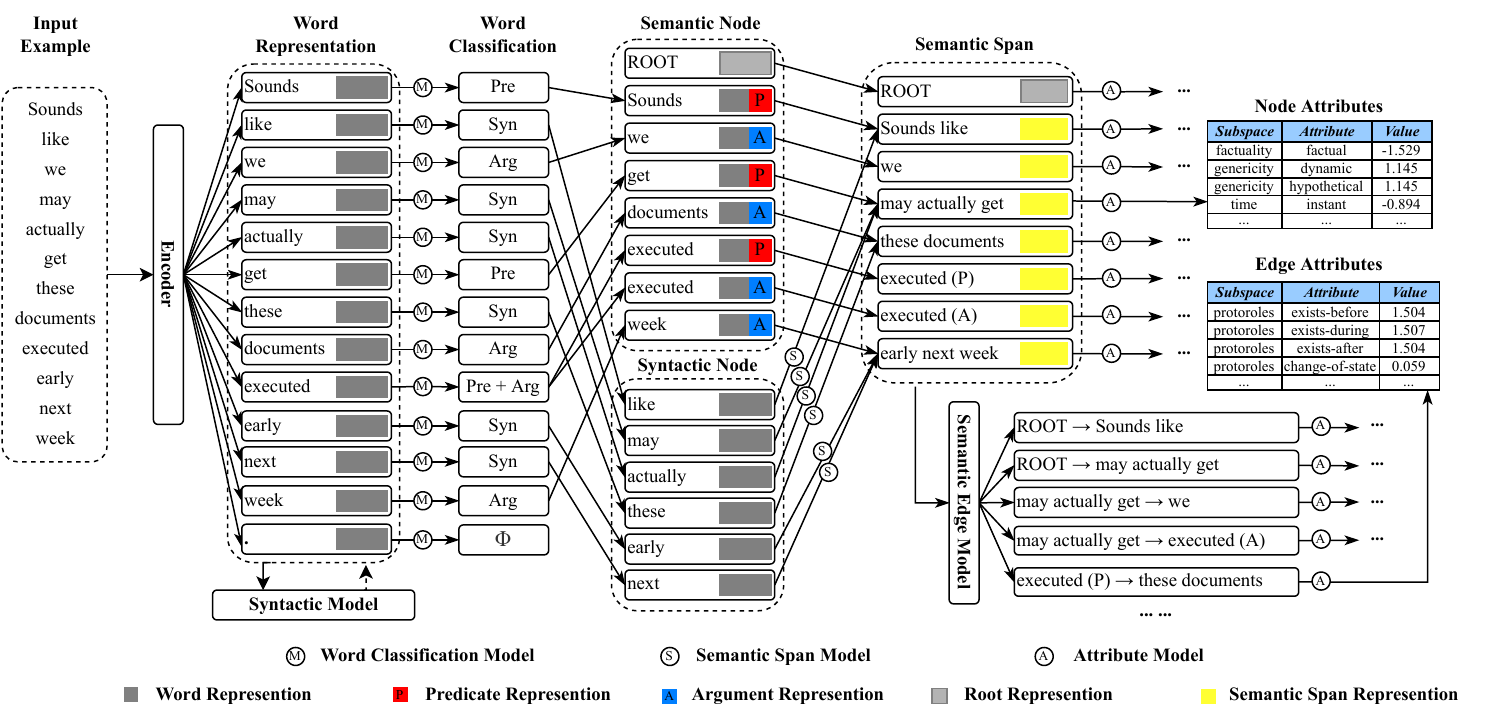}
        \vspace{-8pt}
	\caption{The data flow of our cascade model. The detailed defination of every block are shown in \S\ref{sec:models}. }
	\label{fig:model}
\end{figure*}

\subsection{Efficient Cascade Model}
\label{sec:models}

As discussed in \S\ref{sec:prelim}, our goal is to predict syntactic information (POS tags and syntactic tree), semantic relations (semantic nodes, edges, and spans), and semantic attributes (node- and edge-level) using a single model. To this end, we propose a cascade model to predict all of these information step by step, as illustrated in Figure~\ref{fig:model}. In the following paragraphs, we discuss each component of our model in detail. The sentence is represented as $x_1, x_2, \dots, x_K$, where $x_i$ represents the $i$-th word. The properties of the node are represented as $t$, the properties of the edge as $e$, the $\operatorname{softmax}$ function as $\sigma$, and the $\operatorname{ReLU}$ function as $R$.

\paragraph{Encoder Module} embeds each word $x_i$ into a corresponding context-aware representation $h_i$. We utilize three types of encoders: multi-layer BiLSTM, transformer encoder, and Pre-trained Language Model (BERT, \citealp{BERTPretrainingDeep_2019}). For BiLSTM and transformer encoder, we employ a similar embedding layer with \citet{JointUniversalSyntactic_2021} to ensure comparability, concatenating GloVe word embeddings \citep{GloveGlobalVectors_2014}, character CNN embeddings, and BERT contextual embeddings. For BERT encoder, we use the default subword embeddings layer and mean-pool over all subwords of a word to obtain the word-level representations.


\paragraph{Syntactic Module} predicts the part-of-speech (POS) tag and the syntactic tree. For POS, we use a simple multi-layer perceptron ($\operatorname{MLP}$) over each word representation $h_i$. For the syntactic tree, each word has exactly one syntactic head, so we predict the headword $x_i^y$ and the corresponding edge type $t_i^y$ for each word $x_i$. We follow the approach of \citet{DeepBiaffineAttention_2017a} and \citet{BroadCoverageSemanticParsing_2019} to use a biaffine parser, formally:
\begin{equation}
\begin{aligned}
    \hat{x}_i^y = & p(x|x_i) \\
    = & \sigma(\operatorname{Biaffine}(R(w_l^yh_i), R(w_r^yh_{1:i-1}))) \\
    \hat{t}_i^y = & p(t_y|x_i, \hat{x}^h_i) \\
    = & \sigma(\operatorname{Bilinear}(R(w_l^th_i), R(w_r^t\hat{h}_{1:i-1}^y)))
    \label{eq:synedge}
\end{aligned}
\end{equation}
where $x \in \{x_1, x_2, \dots, x_K\}$, and $\hat{h}_i^y$ is the representation of the predicted head $\hat{x}_i^y$.

\paragraph{Word Classification Module} predicts the semantic edge directly connected to each word (instance, non-head) and the type of the parent node. We simplify this edge prediction problem into a classification problem. We define:
\begin{compactitem}
\item Type ``$\Phi$'': Words with no connecting edge;
\item Type ``Syn'': Words connect with a non-head edge;
\item Type ``Pre'': Words connect with an instance edge, and its parent is a predicate node;
\item Type ``Arg'': Words connect with an instance edge, and its parent is an argument node;
\item Type ``Pre + Arg'': Words connect with two instance edge, and its parents are a predicate node and an argument node;
\end{compactitem}
We use a simple $\operatorname{MLP}$ for classification, formally:
\begin{equation}
    \hat{t}_i^m = p(t_m | x_i) = \sigma(\operatorname{MLP}(h_i))
    \label{eq:semnode}
\end{equation}


\paragraph{Node Generation} predicts the syntactic nodes, semantic nodes, and their corresponding node embeddings. Syntactic nodes $n_1,$ $n_2, \dots, n_N$ have label ``Syn'', and we define the embedding $g^n_i$ of the syntactic node $n_i$ the same as its word embeddings. semantic nodes $m_1,$ $m_2, \dots, m_M$ have label ``Pre'', ``Arg'', or ``Pre + Arg''. We generate will two nodes for ``Pre + Arg''. So for node embeddings, we first concatenate a node type embedding with its word embedding to distinguish whether it is an argument or predicate node. Then we project it back to the previous dimension with a linear layer to generate the embedding $g^m_i$ of the semantic node $m_i$. Furthermore, we generate a virtual root node for every sentence with the same trainable embeddings.

\paragraph{Semantic Span Module} predicts the semantic span by separating each syntactic node to the semantic nodes. Each syntactic node belongs to exactly one semantic node. So we use the same model as Eq.~\ref{eq:synedge} to predicted which semantic node $m_i^h$ is the syntactic node $n_i$ belongs to, formally:
\begin{equation}
\begin{aligned}
    \hat{m}_i^h = & p(m|n_i)\\
    = & \sigma(\operatorname{Biaffine}(R(w_l^mg^n_i), R(w_r^mg_{1:i-1}^m)))
    \label{eq:semspan}
\end{aligned}
\end{equation}
where $m \in \{m_1, m_2,\dots,$ $m_M\}$. The new span level embedding $g^s_i$ for the semantic node $m_i$ is the same as $g^m_i$ by default. Besides, we have also tried to refine $g^s_i$ with the syntactic node embedding, which does not achieve obvious effects.

\paragraph{Semantic Edge Module} predicts the edge and the corresponding type $e_{i,j}^m$ between any two semantic nodes. We consider the case where there is no edge between two semantic nodes as a special type $\Phi$. For prediction, we consider the span level embedding for each pair of nodes, formalized as:
\begin{equation}
\begin{aligned}
    \hat{e}_{i,j}^m = & p(e_m|m_i, m_j)\\
    = & \sigma(\operatorname{Biaffine}(R(w_l^eg^s_i), R(w_r^eg^s_j)))
    \label{eq:semedge}
\end{aligned}
\end{equation}

\paragraph{Attribute Module} predicts the node-level attributes $\hat{t}_i^{a}$ for node $m_i$, and edge-level attributes $\hat{e}_{i,j}^{a}$ for edge between $m_i$ and $m_j$. We use the $\operatorname{MLP}$ model as the main part, formalized as follows:
\begin{equation}
\begin{aligned}
    \hat{t}_i^{a} = & \operatorname{MLP}(g^s_i) \\
    v_i= & R(w_l^vg^s_i), \; v_j=R(w_r^vg^s_j) \\
    \hat{e}_{i,j}^{a} = & \operatorname{MLP}([v_i^TWv_j, v_i, v_j])
    \label{eq:attr}
\end{aligned}
\end{equation}
Here, $W \in \mathbb{R}^{d_v \times d_v \times d_o}$, where $d_v$ is the dimension of $v_i$ and $v_j$, and $d_o$ is the output dimension. $i,j$ must satisfy $\hat{e}_{i,j}^m \neq \phi$ for $\hat{e}_{i,j}^{a}$ (edge exists). Note that attributes may not exist, and we use the same model as above to predict the mask of attributes.

\paragraph{Loss} To train our models, we use different loss functions depending on the task. For word classification, semantic span, and semantic edge modules, we use cross-entropy loss. For attribute module, when predicting the mask, we use binary cross-entropy loss. When predicting the attribute, we follow \citet{UniversalDecompositionalSemantic_2020} to use a composite loss function $\mathcal{L}$ for the values, formally:
\begin{equation}
\small
    \mathcal{L}_{attr}^{value}\left(\hat{t}, t\right)= \frac{2 \cdot \mathcal{L}_{\mathrm{MSE}}\left(\hat{t}, t\right) \cdot \mathcal{L}_{\mathrm{BCE}}\left(\hat{t}, t\right)}{\mathcal{L}_{\mathrm{MSE}}\left(\hat{t}, t\right)+\mathcal{L}_{\mathrm{BCE}}\left(\hat{t}, t\right)}
    \label{eq:attrloss}
\end{equation}
where $\mathcal{L}_{\mathrm{MSE}}$ is the mean squared loss, $\mathcal{L}_{\mathrm{BCE}}$ is the binary cross-entropy loss, $t$ is the gold attribute, and $\hat{t}$ is our prediction. $\mathcal{L}_{\operatorname{MSE}}$ encourages the predicted attribute value to be close to the true value, while $\mathcal{L}_{\operatorname{BCE}}$ encourages the predicted and reference values to share the same sign. 

Finally, to handle this multi-task problem, we use a weighted sum of all the loss functions mentioned above for our model:
\begin{equation}
\small
\mathcal{L}=a_1\mathcal{L}_{cls} + a_2\mathcal{L}_{span} + a_3\mathcal{L}_{edge} + a_4\mathcal{L}_{attr}^{mask} + a_5\mathcal{L}_{attr}^{value}
    \label{eq:totloss}
\end{equation}
where $a_i = 1$ for $i \in [1, .., 5]$, except $a_2 = 2$.

\subsection{Incorporating Syntactic Information}
\label{sec:addsyn}

By default, we incorporate syntactic information by \emph{multi-task training}. Additionally, we propose \emph{GCN} and \emph{attention} approaches for a more profound incorporation of syntactic information. Specifically, we utilize the syntactic information to update the word embeddings generated by the encoder. The strategies are as follows:

\paragraph{Multi-task Training}
We add the loss of the syntactic module to term $\mathcal{L}$, which incorporates syntactic information into the shared encoder through back-propagation. We use cross-entropy loss for POS and syntactic tree parsing, formally:
\vspace{-3pt}
\begin{equation}
\small
\mathcal{L_{\text{+}SYN}}=\mathcal{L} + a_6\mathcal{L}_{pos} + a_7\mathcal{L}_{tree}
\vspace{-3pt}
\label{eq:synloss}
\end{equation}
where $a_6 = a_7 = 1$.

\paragraph{GCN} Inspired by the idea of GCN \citep{SemiSupervisedClassificationGraph_2017a}, we try to encode the predicted adjacency matrix information into the embedding. In the syntactic tree, we consider two types of edges: directed edges from parent nodes (top) to child nodes (bottom), and those with reverse directions. Then we employ a bidirectional GCN consisting of 1) top-down GCN to convey sentence-level information to local words, and 2) bottom-up GCN to convey phrase-level information to the center word. Additionally, to further convey the edge type information corresponding to the current word, we 3) consider the probability distribution of its edge type, and use a GCN-like method to convey this information. With the word embedding matrix $\mathbf{H}$ being the input $\mathbf{H}^{(0)}$, we use a $l$ layer model (with $l=2$ in practice), formally:
\begin{equation}
\begin{aligned}
    \mathbf{V}^{(i)} = & [\mathbf{A}_h\mathbf{H}^{(i)}\mathbf{W}_1^{(i)}, \mathbf{A}_h^T\mathbf{H}^{(i)}\mathbf{W}_2^{(i)}, \mathbf{A}_t\mathbf{T}_{e}\mathbf{W}_3^{(i)}] \\
    & \mathbf{H}^{(i+1)} = R(\mathbf{W}_4^{(i)}R(\mathbf{V}^{(i)})) \\
    & \mathbf{H}_o = \mathbf{W}_o[\mathbf{H}^{(0)}, \mathbf{H}^{(l)}]
    \label{eq:udgcn}
\end{aligned}
\end{equation}
where $\mathbf{A}_h$ is the top-down adjency matrix prediction, $\mathbf{A}_h^T$ is the bottom-up ones, $\mathbf{A}_t$ is the edge type probability distribution, and $\mathbf{T}_{e}$ is the trainable edge type embedding matrix. Note that the adjacency matrix does not self-loop, so GCN does not convey information about the words themselves. We then combine the original word embeddings $\mathbf{H}^{(0)}$ with the output $\mathbf{H}^{(l)}$ to get the new word embeddings $\mathbf{H}_o$. Under such a design, good results can be achieved with a relatively shallow network.

\paragraph{Attention} Word representation after dimension reduction used in syntactic edge and type prediction contains basic information of the syntactic tree \citep{JointUniversalSyntactic_2021}. So we directly use the representations in Eq.~\ref{eq:synedge}, which are used in the $\operatorname{Biaffine}$ and $\operatorname{Bilinear}$ model. Formally:
\begin{equation}
\begin{aligned}
    \mathbf{V} = & R([w_l^w\mathbf{H}, w_r^w\mathbf{H}, w_l^t\mathbf{H}, w_r^t\mathbf{H}]) \\
    & \mathbf{H}_o = \mathbf{W}_o[\mathbf{H}, \mathbf{A}_h \mathbf{V}]
    \label{eq:udctat}
\end{aligned}
\end{equation}
where all the $w_*^*\mathbf{H}$ come from Eq.~\ref{eq:synedge} without recalculation. Comparing to the GCN approach, this method uses fewer new parameters and requires less additional calculation, while still preserving the performance improvements achieved by the GCN approach to some extent.


\begin{figure*}[t]
    \centering
    \includegraphics[width=0.88\textwidth]{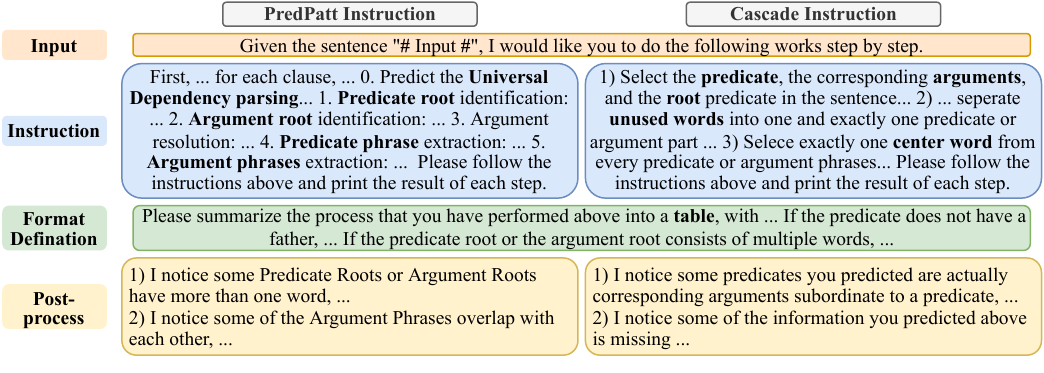}
    \vspace{-12pt}
    \caption{The prompt for semantic relation parsing for ChatGPT. For each generation, we first input the input and instruction, then input the format definition to get the prediction. Finally, we input the post-process part one by one to generate better predictions, i.e., three predictions for a sentence in a single conversation.}
    \vspace{-6pt}
    \label{fig:prompt}
\end{figure*}

\subsection{Data Augmentation}
\label{sec:data-centric}

One of the features of UDS dataset is the strong correlation with external tools PredPatt. \citet{UniversalDecompositionalSemantic_2020} attempt to use an external model to predict the POS tags and syntactic tree on the test dataset, which are then fed directly to PredPatt to obtain the semantic relations. However, the effectiveness of this method is relatively poor, likely due to two issues: 1) the error transmission problem comes from the prediction of the syntactic model, and 2) the rule-based tools are not as robust as neural networks towards noisy inputs.

To address these issues, we propose a data augmentation method. Instead of using it during inference, we use it to augment the data, with only the help of external unlabeled data. Specifically, we first train a model to predict the syntactic tree and POS tag, using the above syntactic model. Next, we use PredPatt to generate pseudo labels (i.e., semantic relations) for the unlabeled data. Finally, we use these data to pre-train our model, and then fine-tune it with a smaller learning rate using the labeled UDS dataset, which achieves significant improvements in relation parsing.

\subsection{Large Language Models}
\label{sec:llm}
We explore the direct use of large language models (LLMs) for parsing tasks. First, for semantic relation parsing, we try two types of prompts shown in Figure~\ref{fig:prompt}: 1) The Predpatt prompt guides the LLM to first generate the syntactic parsing, then follow the instruction of the PredPatt tool step by step to generate the semantic relations. 2) Cascade approach follows the idea of our model to decompose the UDS parsing, which first selects the center phrase of the semantic node, then expands every phrase into a span. To make sure that the center word has only one word, we select it at the last step. Second, for semantic attribute parsing, we provide the sentence and the corresponding node/edge as input, definitions of attribute types as instruction, and conduct experiments under the oracle setting defined in \S\ref{sec:metric}. As the scale of attribute scoring may vary across conversation rounds, we only let it predict positive or negative.

Additionally, we investigate the generation of new data for downstream model training, which is widely used. We use the random token lists as input rather than the unlabeled data, and let LLM generate the POS tag and syntax tree, which are further used to generate pseudo-labels, following \S\ref{sec:data-centric}. Since Universal Dependencies is a widely used dataset and contains both of the required information, a simple prompt can be used.



\begin{table*}[t]
\small
\centering
\begin{tabular}{clC{1.15cm}C{1.15cm}C{1.15cm}C{1.15cm}cC{1.15cm}C{1.15cm}C{1.15cm}}
\toprule
~ & \bf Strategy & \bf S-P & \bf S-R & \bf S-F1 & \bf Attr. $\mathbf{\rho}$ & \bf Attr. F1 & \bf UAS & \bf LAS & \bf POS \\ \midrule
\multirow{4}*{\rotatebox{90}{\small \textbf{Baseline}}} & LSTM & 89.90 & 85.85 & 87.83 & 0.46 & 60.41 & - & - & - \\
~ & \enspace + SYN & 88.58 & 87.67 & 88.12 & 0.46 & 61.28 & 91.44 & 88.80 & - \\
~ & TFMR  & 90.04 & 87.98 & 89.19 & 0.56 & 67.89 & - & - & - \\
~ & \enspace + SYN  & \bf 91.09 & 89.01 & 90.04 & 0.56 & 66.85 & 92.40 & 89.96 & - \\ \midrule
\multirow{9}*{\rotatebox{90}{\small \textbf{Mine}}} & LSTM & 87.75 & 91.12\dag & 89.79\dag & 0.47\dag & 57.93 & - & - & - \\
~ & \enspace + SYN & 88.82\dag & 92.50\dag & 90.62\dag & 0.46 & 57.34 & 91.71 & 89.10 & 96.29 \\
~ & \enspace + SYN + DA & 90.00 & 93.37 & 91.65 & 0.33 & 49.66 & 92.65 & 90.51 & 96.67 \\
~ & TFMR & 88.34 & 92.90\dag & 90.56\dag & 0.49 & 59.68 & - & - & - \\
~ & \enspace + SYN & 89.28 & 93.56\dag & 91.37\dag & 0.49 & 58.45 & 92.07 & 89.65 & 96.85 \\
~ & \enspace + SYN + DA & 90.15 & 93.49 & 91.79 & 0.42 & 54.27 & \bf 93.03 & 90.91 & 97.10 \\
~ & BERT & 88.90 & 92.77\dag & 90.79\dag & \bf 0.60\dag & 67.02 & - & - & - \\
~ & \enspace + SYN & 89.51 & 94.18\dag & 91.79\dag & 0.59\dag & 65.78 & 92.81\dag & 90.73\dag & \bf 97.18 \\
~ & \enspace + SYN + DA & 90.27 & \bf 94.23 & \bf 92.20 & 0.54 & 63.91 & 92.98 & \bf 90.93 & 97.08 \\ \midrule
\multirow{3}*{\rotatebox{90}{\small \textbf{LLM}}} & PRED & 35.50 & 51.28 & 41.96 & - & - & - & - & - \\
& CASC & 38.13 & 53.26 & 44.44 & - & - & - & - & - \\
& ATTR & - & - & - & - & \bf 80.69 & - & - & - \\
\bottomrule
\end{tabular}
\caption{\label{tab:casmodel}
Main results. ``LSTM'', ``TFMR''(Transformer), ``BERT'' stands for different encoder. We run t-test against the corresponding baseline, and $\dag$ means significantly higher with $>95\%$ confidence. ``+SYN'' means GCN approach in \S\ref{sec:addsyn}, and ``DA'' means the data augmentation method in \S\ref{sec:data-centric}. ``PRED'' and ``CASC'' are Predpatt and Cascade Instruction for semantic relation parsing, respectively, and ``ATTR'' is the semantic attribute parsing, detailed in \S\ref{sec:llm}. Metric abbreviation are detailed in \S\ref{sec:metric}.}
\end{table*}

\section{Experiment and Analysis}

\subsection{Experimental Setup}
\label{sec:expsetup}

\paragraph{Datasets} We conduct experiments on the UDS dataset \citep{UniversalDecompositionalSemantics_2020}, with 10k valid training sentences. For English monolingual data, we use publicly available News Crawl 2021 corpus \citep{ExploitingSourcesideMonolingual_2016, ExploitingMonolingualData_2019a}. In the experiment of the data augmentation method, we first generate the pseudo-targets for all the monolingual data, then filter out the ones that have invalid syntactic and semantic graphs. Finally we randomly select a 100k corpus subset. For ChatGPT generation, we select a 10k corpus subset.

\paragraph{Model Training} Our model is trained on one NVIDIA A30 Tensor Core GPU with a batch size of 16 and a dropout rate of 0.3. We fix BERT parameters for LSTM and transformer encoders and keep them trainable when BERT itself is the encoder. For the majority of the training process, we set the learning-rate to 2e-4, while for BERT encoder, we set it to 1e-5. For a fair comparison, we use a linear projection of the output of all the encoders to unify the output dimension to 1024. We run each model five times under different seeds in the main table and show the average score. For ChatGPT, we build experiments in the dialog box, using the ChatGPT Mar 23 Version.

\paragraph{Baseline} We use \citet{JointUniversalSyntactic_2021} as the baseline. It first employs GloVe word embeddings \citep{GloveGlobalVectors_2014}, character CNN embeddings, and BERT \citep{BERTPretrainingDeep_2019} to generate the context-aware representations of the input sentence. Then, it generates each edge with a decoder in an autoregressive way, following the idea of a pointer-generator network \citep{GetPointSummarization_2017}. After that, it uses a deep biaffine \citep{DeepBiaffineAttention_2017a} graph-based parser to create edges. Node- and edge-level attributes are then predicted after every step, with a multi-layer perception for node attributes and a deep biaffine for edge attributes. Besides, the introduction of syntactic information is preliminarily tried, and we only report their optimal results for each metric.

\subsection{Metrics}
\label{sec:metric}

We follow the setting given by \citet{JointUniversalSyntactic_2021}, detailed as follows.

\paragraph{S-score} 
This metric measures performance on the semantic relation prediction task. Following the Smatch metric \citep{SmatchEvaluationMetric_2013a}, which uses a hill-climbing approach to find an approximate graph matching between a reference and predicted graph, S-score \citep{EvaluationPredPattOpen_2017a} provides precision (S-P), recall (S-R), and F1 score (S-F1) for nodes, edges, and attributes. We follow \citet{JointUniversalSyntactic_2021} and evaluate the S-score for nodes and edges only, which evaluates against full UDS arborescences with linearized syntactic subtrees included as children of semantic heads.

\paragraph{Attribute $\mathbf{\rho}$ \& F1} For UDS attributes, we use the pearson correlation $\rho$ (Attr. $\rho$) between the predicted attributes at each node and the gold annotations in the UDS corpus. We also use F1-score (Attr. F1) to measure whether the direction of the attributes matches that of the gold annotations. We binarized the attribute with threshold value $\theta=0$ for gold attributes, and tune $\theta$ for predicted ones per attribute type on validation data. Both of them are obtained under an ``oracle'' setting, where the gold graph structure is provided.

\paragraph{Syntactic Metric} We follow \citet{DependencyParsingMetrics_2015a} to use Unlabeled Attachment Score (UAS) to compute the fraction of words with correctly assigned heads, and Labeled Attachment Score (LAS) to compute the fraction with correct heads and edge types. While for part-of-speech (POS), we simply use the accuracy of prediction.

\subsection{Main Results}
\label{sec:expmain}

We conduct experiments on three types of encoders, as demonstrated in Table~\ref{tab:casmodel}.

\begin{figure}[t]
    \centering
    \includegraphics[width=0.38\textwidth]{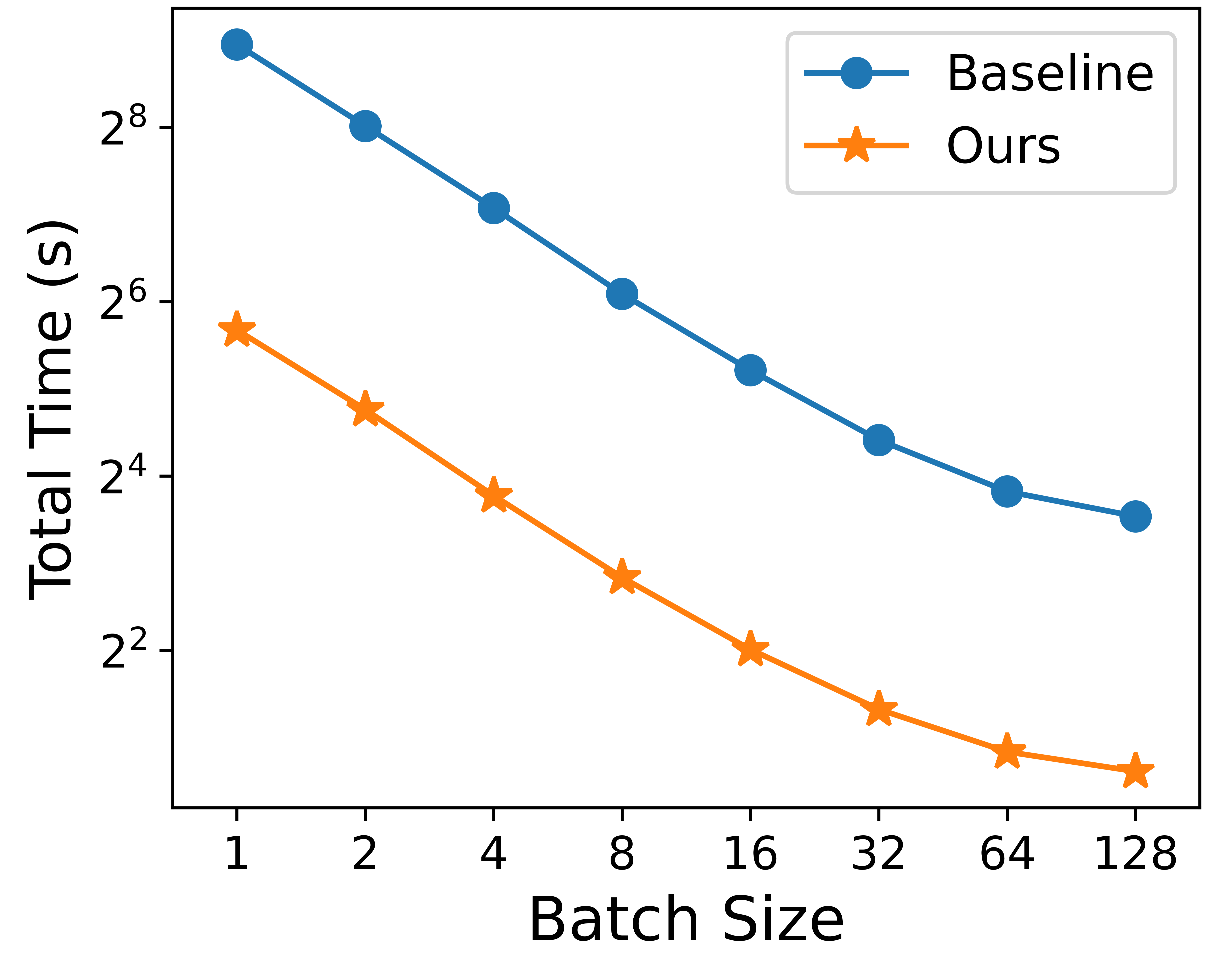}
    \vspace{-12pt}
    \caption{Total inference time for forward propagation of the two models, varying with the batch size. We use logarithmic coordinates for better comparison.}
    \label{fig:time}
\end{figure}

\begin{figure}[t]
    \centering
    \includegraphics[width=0.38\textwidth]{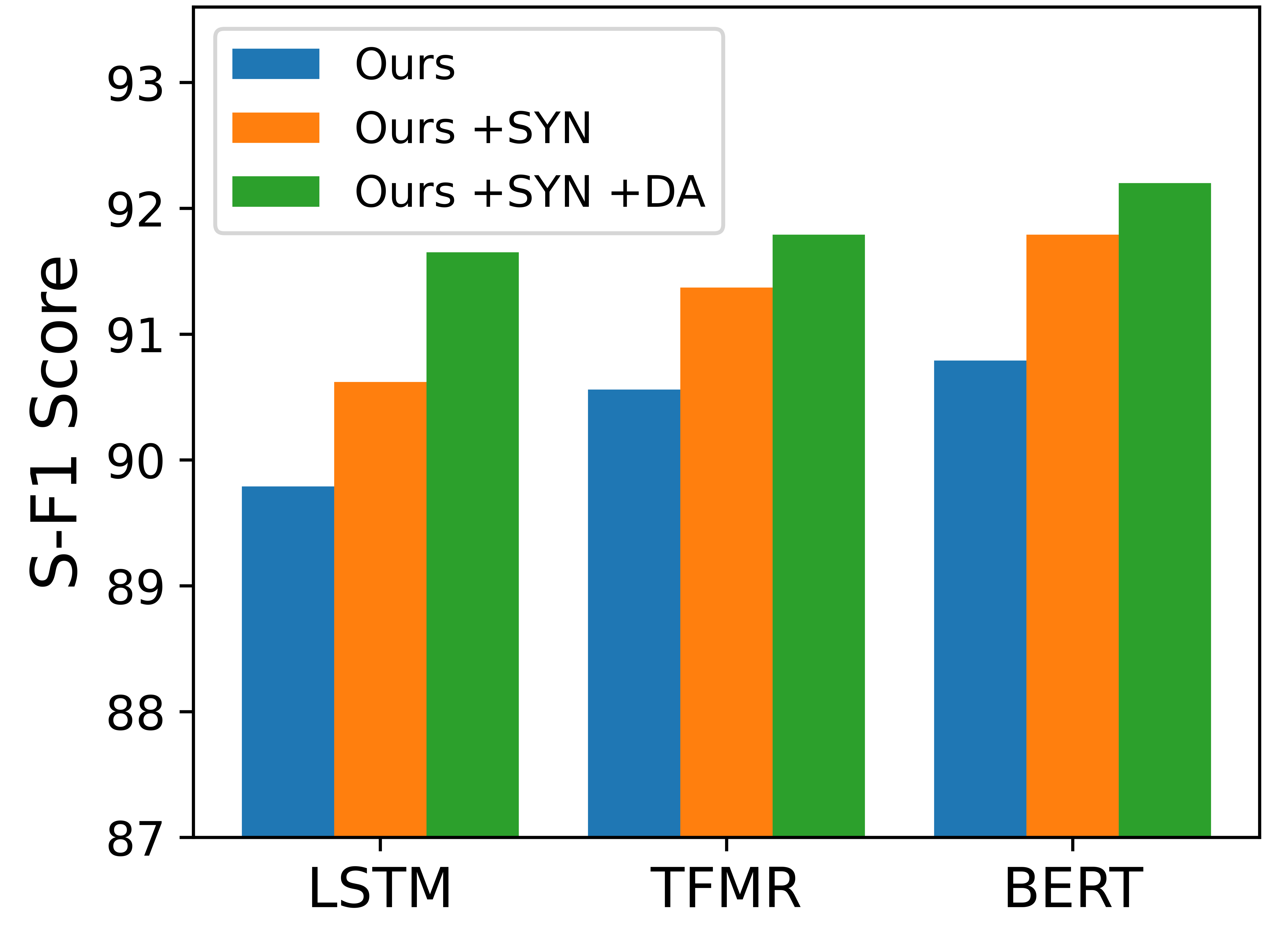}
    \vspace{-12pt}
    \caption{S-F1 score of different encoders. Abbreviations are defined in Table~\ref{tab:casmodel}.}
    \label{fig:semsf}
\end{figure}

\paragraph{Our cascade model outperforms the baseline model.} Under basic settings, our best setting (BERT) significantly improves the baseline (TFMR) in S-F1 (+1.60) and Attr. $\rho$ (+0.04), and slightly worse in Attr. F1. The above results are also preserved under +SYN settings (+1.75 and +0.03, respectively). Furthermore, we calculated the total inference time for forward propagation of the two models, averaging on validation and test datasets (about 1.3k sentences). The results are shown in Figure~\ref{fig:time} under logarithmic coordinates. Our model significantly reduces the inference time for all batch sizes (9.56 times faster on average). Finally, using additional data augmentation methods, the S-F1 can be further improved (+2.16), which is also held in LSTM and Transformer (+3.53 and +1.75, respectively). The above results show that our model significantly outperforms the baseline.


\paragraph{Syntactic information and data augmentation methods enhance semantic relation parsing.} Our model primarily focuses on improving the semantic relation parsing, which LLMs are not good at. We summarize the corresponding result in Figure~\ref{fig:semsf}. We can see that both the two approaches can significantly improve relation parsing, with +0.88 for syntactic information and +0.62 for the data augmentation method on average. Besides, the improvements are orthogonal to each other and can be used simultaneously, pushing the results of different models towards a similar limit, since lower-performing models experience greater improvements.

\paragraph{The same methods do not benefit attribute prediction.} However, our proposed methods for further improvements do not consistently improve the attribute parsing. Attributes derive from crowdsourced annotation, which is not closely related to the syntactic or semantic information. Thus, syntactic information cannot provide useful information for attribute prediction, and using more data to pre-train a better model for semantic relation parsing is harmful to the performance of attribute parsing.

\begin{figure*}[t]
    \centering
    \includegraphics[width=0.97\textwidth]{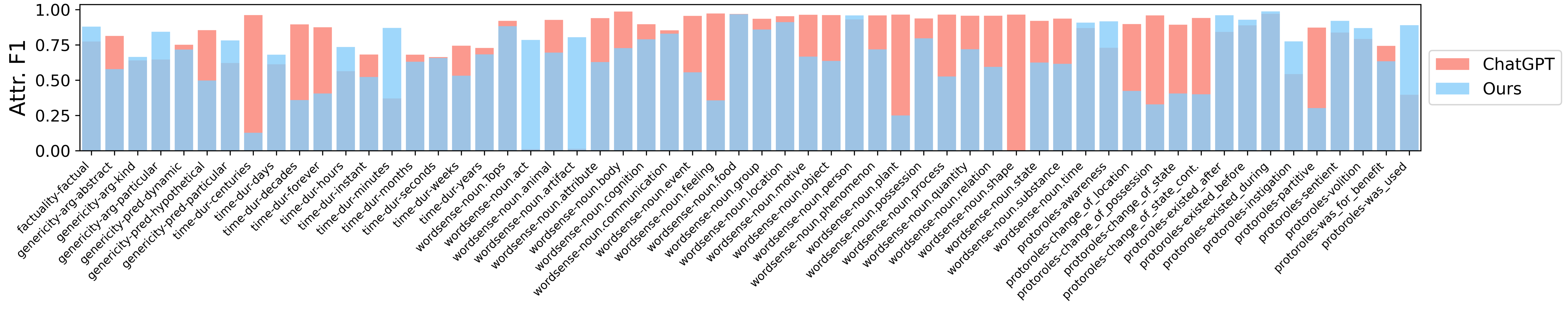}
    \vspace{-12pt}
    \caption{S-F1 score for each UDS attribute using ChatGPT (80.69 on average) or using our basic BERT model (67.02 on average). The x-axis is the UDS attribute name, with the ones beginning with ``protoroles'' being the edge-level attributes (the rightmost 14 attributes), and other attributes are at the node-level.}
    \label{fig:chatgpt}
\end{figure*}

\begin{table*}[t]
\small
\centering
\begin{tabular}{clC{1.1cm}C{1.1cm}C{1.1cm}C{1.1cm}cC{1.1cm}C{1.1cm}C{1.1cm}}
\toprule
~ & \bf Strategy & \bf S-P & \bf S-R & \bf S-F1 & \bf Attr. $\mathbf{\rho}$ & \bf Attr. F1 & \bf UAS & \bf LAS & \bf POS \\ \midrule
\multirow{5}*{\rotatebox{90}{\small \textbf{LSTM}}} & Naive & 87.75 & 91.12 & 89.79 & 0.47 & 57.93 & - & - & - \\
& \enspace + Joint & 88.21 & 92.51 & 90.31 & 0.45 & 55.42 & 91.51 & 89.07 & 96.23 \\
~ & \enspace + Attn. & 88.62 & 92.55 & 90.54 & 0.45 & 57.65 & 91.95 & 89.41 & 96.26 \\
~ & \enspace + GCN & 88.82 & 92.50 & 90.62 & 0.46 & 57.34 & 91.71 & 89.10 & 96.29 \\
~ & \enspace + Span & 88.45 & 92.31 & 90.34 & 0.47 & 57.85 & 91.58 & 89.08 & 96.37 \\ \midrule
\multirow{5}*{\rotatebox{90}{\small \textbf{TFMR}}} & Naive & 88.34 & 92.90 & 90.56 & 0.49 & 59.68 & - & - & - \\
~ & \enspace + Joint & 88.64 & 93.53 & 91.02 & 0.51 & 59.56 & 91.99 & 89.42 & 96.60 \\
~ & \enspace + Attn. & 88.82 & 93.46 & 91.08 & 0.49 & 58.86 & 91.84 & 89.35 & 96.77 \\
~ & \enspace + GCN & 89.28 & 93.56 & 91.37 & 0.49 & 58.45 & 92.07 & 89.65 & 96.85 \\
~ & \enspace + Span & 88.85 & 93.19 & 90.97 & 0.50 & 59.46 & 91.60 & 89.29 & 96.71 \\ \midrule
\multirow{5}*{\rotatebox{90}{\small \textbf{BERT}}} & Naive & 88.90 & 92.77 & 90.79 & 0.60 & 67.02 & - & - & - \\
& \enspace + Joint & 88.87 & 93.75 & 91.25 & 0.60 & 67.63 & 92.95 & 90.79 & 97.12 \\
~ & \enspace + Attn. & 89.25 & 94.05 & 91.59 & 0.58 & 66.60 & 92.94 & 90.77 & 97.02 \\
~ & \enspace + GCN & 89.51 & 94.18 & 91.79 & 0.59 & 65.78 & 92.81 & 90.73 & 97.18 \\
~ & \enspace + Span & 88.95 & 93.64 & 91.23 & 0.59 & 65.97 & 92.92 & 90.74 & 97.23 \\
\bottomrule
\end{tabular}
\caption{\label{tab:combinsyn}
The effect of different strategies to incorporate syntactic information. ``Naive'' means no additional syntactic information. ``+Joint'', ``+Attn'', and ``+GCN'' means incorporating syntactic information using joint training, GCN, and attention in \S\ref{sec:addsyn}, separately. ``+Span'' means refine span embeddings using syntactic nodes.}
\end{table*}

\paragraph{ChatGPT performs poorly on relation parsing.} For semantic relation parsing, we use the prompt given in \S\ref{sec:llm} 3 times, which generates 9 different results. We filter out invalid output (no table or table with incorrect headers) and select the best result for each sentence. There are still 11.04\% and 0.37\% of the sentences that do not have correct results for PRED and CASC, respectively, which are filtered out. Despite this favorable setting, it still achieved poor results. Under our observation, the generated relations of LLMs are typically semantically compliant. However, they struggle to follow the instruction step by step, leading to outputs that often do not meet our requirements, and repetitions and incorrect summarizations in the table also commonly occur. As a result, LLMs perform poorly on relation parsing, especially in precision, and complex post-processing constructed by professionals is highly required.

\paragraph{ChatGPT performs perfectly on attribute parsing.} For semantic attribute parsing, we only run ChatGPT once. 3.03\% of the sentences do not have correct results and are filtered out. Results show that ChatGPT significantly outperforms the small models, achieving a +12.80 increase in Attribute F1 scores compared to the best model. We think that for ChatGPT, which is well-aligned with humans, it is easier to predict the attributes given by human annotators rather than the long logical chain reasoning task. In addition, only need to predict positive and negative without considering the pearson correlation is also one of its advantages. For further verification, we calculate the Attribute F1 scores for all attributes in Figure~\ref{fig:chatgpt}. We can observe that ChatGPT performs well on most of the attributes when compared to our model, with 60.34\% and 25.86\% of the attributes respectively having F1 scores above 85\%. Furthermore, ChatGPT performs perfectly on word-sense attributes, achieving an F1 score of 86.99. In contrast, our models do not display significantly superior results, with an F1 score of 68.49. We believe that with more detailed guidance and rigorous post-processing, LLMs have the potential to replace humans in annotation tasks.

\subsection{Exploration on Syntactic Information}
\label{sec:expsyn}

We conduct experiments on different ways to join syntactic information into the model, and the results are shown in Table~\ref{tab:combinsyn}.

\paragraph{Syntactic information enhances semantic relation parsing.} Our experiments show consistent improvements in S-F1 scores across different methods of integrating syntactic information, with +0.48 for +SYN, +0.69 for Contact, and +0.88 for GCN, which is also used as our default settings. However, because of the different syntactic foundations arising from different annotation methods, we do not observe a consistent trend of attribute parsing results, aligned with the findings in \citet{JointUniversalSyntactic_2021}. 

\paragraph{Incorporating child syntactic information has less impact on the results.} We tried to use a better span representation, which uses a self-attention over all words in the span, instead of using only the representation of the center word. However, the attribute prediction does not achieve consistent improvements. This shows that the center word can well represent the semantics of the whole span, and is the default setting in our experiments.

\begin{table*}[t]
\small
\centering
\begin{tabular}{lcccC{1.05cm}C{1.05cm}C{1.05cm}C{1.0cm}C{1.05cm}C{1.05cm}C{1.05cm}}
\toprule
\bf Model & \bf In domain & \bf Predpatt & \bf S-P & \bf S-R & \bf S-F1 & $\bf \Delta$ & \bf UAS & \bf LAS & \bf POS \\ \midrule
Ours & $\times$ & $\times$ & 89.34 & 94.05 & 91.63 & +0.38 & 93.05 & 91.01 & 97.10 \\
Ours & $\times$ & $\checkmark$ & 90.15 & 94.28 & 92.17 & +0.92 & 93.26 & 91.28 & 97.17 \\
Ours & $\checkmark$ & $\times$ & 89.08 & 94.03 & 91.49 & +0.24 & 92.62 & 90.52 & 97.19 \\
Ours & $\checkmark$ & $\checkmark$ & 89.56 & 94.37 & 91.90 & +0.65 & 92.88 & 90.91 & 97.22 \\
Stanza & $\times$ & $\checkmark$ & 89.24 & 94.12 & 91.62 & +0.37 & 92.96 & 90.86 & 97.27 \\
Stanza & $\checkmark$ & $\checkmark$ & 89.69 & 94.24 & 91.90 & +0.65 & 92.76 & 90.74 & 97.10 \\
ChatGPT & $\times$ & $\checkmark$ & 88.25 & 93.28 & 90.70 & -0.55 & 92.38 & 90.28 & 96.99 \\ \midrule
\multicolumn{3}{l}{Syntactic Teacher} & - & - & - & & 93.24 & 91.14 & 97.54 \\
\multicolumn{3}{l}{Semantic Relation Teacher} & 88.87 & 93.75 & 91.25 & - & 92.95 & 90.79 & 97.12 \\
\bottomrule
\end{tabular}
\vspace{-2pt}
\caption{\label{tab:datacentric}
The effect of different data augmentation approaches. ``Model'' means which teacher model to use, ``In domain'' means whether to select data with closer domain, and ``Predpatt'' means whether to use an external tool or simply use the distillation method. ``Syntactic Teacher'' is trained only on syntactic targets, while ``Semantic Relation Teacher'' on syntactic and semantic relation targets. Both only use multi-task learning methods.}
\vspace{-8pt}
\end{table*}

\subsection{Exploration on Data Augmentation}
\label{sec:expdata}

We conduct experiments using the data augmentation method under the basic multi-task training method to incorporate syntactic information, and the results are shown in Table~\ref{tab:datacentric}.

\paragraph{Data augmentation significantly improves the semantic parsing.} Under different ways to incorporate syntactic information, the S-F1 consistently improves, with +0.54 on average and +0.92 for best settings (ours w/o in domain w/ predpatt), which is used as the default data augmentation method. Besides, our proposed ways to better utilized the external tool also significantly outperforms the basic distillation settings, i.e., +0.48 on average, which shows the efficacy of our methods.

\paragraph{How does the in-domain unlabeled-data act?} We are also curious about how the domain of the datasets influences the results. We follow the idea of \citet{IntelligentSelectionLanguage_2010a} to score the unlabeled data by the difference between the score of the in-domain language model and the language model trained from which the unlabeled data is drawn. We refer the reader to the original paper for further details. Results have shown that for our larger models with better generalization, the in-domain data hurt the performance (-0.27). For smaller model given in Stanza, the in-domain data performs better (+0.28), while both are worse than the results with our models. This shows that the performance of the teacher model is important, and for models with good generalization, always using in-domain data is not a good choice.
\vspace{12pt}
\begin{figure}[t]
    \centering
    \includegraphics[width=0.36\textwidth]{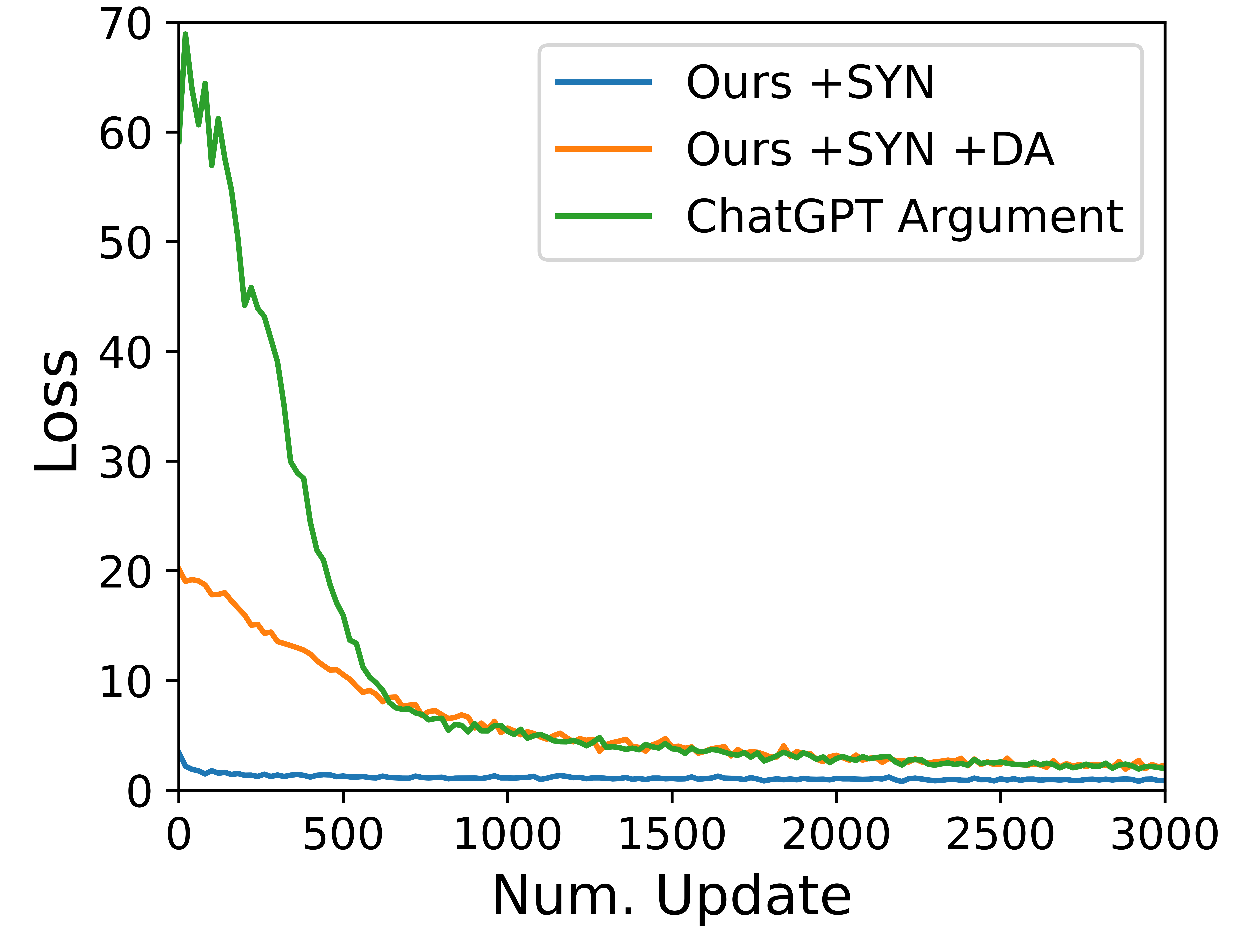}
    \vspace{-12pt}
    \caption{Loss function curve over the first 3k steps.}
    \vspace{-12pt}
    \label{fig:loss}
\end{figure}

\paragraph{How does zero-shot setting with ChatGPT act?} LLMs do not perform well in semantic relation generation, and directly using external tools to assist the test set is not effective \citep{UniversalDecompositionalSemantic_2020}, so it is natural to think of using LLM to augment the data in a similar way. We used the zero-shot settings, detailed in \S\ref{sec:llm}. However, the performance has declined. For further analysis, we propose the training loss for the first 3k updates for different models in Figure~\ref{fig:loss}. We can see that our data augmentation method can significantly lower the initial training loss, which shows that similar data distribution is shared between our proposed pseudo-labeled data and the training data. However, the initial loss of ChatGPT argumentation is even higher than random initializing (Ours +SYN). This shows significant distribution shifts for the data generated by ChatGPT, which shows earge need for more detailed prompts and ways to select properly generated data.

\section{Conclusion}

In this paper, we conduct a holistic exploration of the Universal Decompositional Semantic (UDS) Parsing. We propose a cascade model and ways to better incorporate syntactic information, which both outperform the baseline, while significantly improving the parallelism and reducing inference time. Data augmentation methods are also detailly explored. Finally, ChatGPT performs poorly in relation parsing and data augmentation, but well in attribute parsing, which shows potential for dataset annotation in place of humans. 

\bibliography{tacl2021v1}

\begin{thebibliography}{40}
\expandafter\ifx\csname natexlab\endcsname\relax\def\natexlab#1{#1}\fi

\bibitem[{Abend and Rappoport(2013)}]{UniversalConceptualCognitive_2013}
Omri Abend and Ari Rappoport. 2013.
\newblock Universal {{conceptual cognitive annotation}} ({{ucca}}).
\newblock In \emph{{{ACL}}}.

\bibitem[{Bahdanau et~al.(2015)Bahdanau, Cho, and
  Bengio}]{NeuralMachineTranslation_2015}
Dzmitry Bahdanau, Kyunghyun Cho, and Yoshua Bengio. 2015.
\newblock Neural {{machine translation}} by {{jointly learning}} to {{align}}
  and {{translate}}.
\newblock In \emph{{{ICLR}}}.

\bibitem[{Banarescu et~al.(2013)Banarescu, Bonial, Cai, Georgescu, Griffitt,
  Hermjakob, Knight, Koehn, Palmer, and
  Schneider}]{AbstractMeaningRepresentation_2013}
Laura Banarescu, Claire Bonial, Shu Cai, Madalina Georgescu, Kira Griffitt, Ulf
  Hermjakob, Kevin Knight, Philipp Koehn, Martha Palmer, and Nathan Schneider.
  2013.
\newblock Abstract {{meaning representation}} for {{sembanking}}.
\newblock In \emph{{{LAW}}@{{ACL}}}.

\bibitem[{Bubeck et~al.(2023)Bubeck, Chandrasekaran, Eldan, Gehrke, Horvitz,
  Kamar, Lee, Lee, Li, Lundberg, Nori, Palangi, Ribeiro, and
  Zhang}]{SparksArtificialGeneral_2023}
S{\'e}bastien Bubeck, Varun Chandrasekaran, Ronen Eldan, Johannes Gehrke, Eric
  Horvitz, Ece Kamar, Peter Lee, Yin~Tat Lee, Yuanzhi Li, Scott Lundberg,
  Harsha Nori, Hamid Palangi, Marco~Tulio Ribeiro, and Yi~Zhang. 2023.
\newblock Sparks of {{artificial general intelligence}}: {{early}} experiments
  with {{gpt-4}}.
\newblock \emph{ArXiv}.

\bibitem[{Cai and Knight(2013)}]{SmatchEvaluationMetric_2013a}
Shu Cai and Kevin Knight. 2013.
\newblock Smatch: an {{evaluation metric}} for {{semantic feature structures}}.
\newblock In \emph{{{ACL}}}.

\bibitem[{Caruana(1997)}]{MultitaskLearning_1997a}
Rich Caruana. 1997.
\newblock Multitask {{Learning}}.
\newblock \emph{Mach. Learn.}, 28(1):41--75.

\bibitem[{Chen and Manning(2014)}]{FastAccurateDependency_2014a}
Danqi Chen and Christopher~D. Manning. 2014.
\newblock A {{fast}} and {{accurate dependency parser}} using {{neural
  networks}}.
\newblock In \emph{{{EMNLP}}}.

\bibitem[{Deng et~al.(2023)Deng, Ding, Liu, Zhang, Tao, and
  Zhang}]{Deng_Ding_Liu_Zhang_Tao_Zhang_2023}
Hexuan Deng, Liang Ding, Xuebo Liu, Meishan Zhang, Dacheng Tao, and Min Zhang.
  2023.
\newblock Improving simultaneous machine translation with monolingual data.
\newblock \emph{AAAI}.

\bibitem[{Devlin et~al.(2019)Devlin, Chang, Lee, and
  Toutanova}]{BERTPretrainingDeep_2019}
Jacob Devlin, Ming-Wei Chang, Kenton Lee, and Kristina Toutanova. 2019.
\newblock {{Bert}}: {{pre-training}} of {{deep bidirectional transformers}} for
  {{language understanding}}.
\newblock In \emph{{{NAACL-HLT}}}.

\bibitem[{Dozat and Manning(2017)}]{DeepBiaffineAttention_2017a}
Timothy Dozat and Christopher~D. Manning. 2017.
\newblock Deep {{biaffine attention}} for {{neural dependency parsing}}.
\newblock In \emph{{{ICLR}}}.

\bibitem[{Govindarajan et~al.(2019)Govindarajan, Durme, and
  White}]{DecomposingGeneralizationModels_2019a}
Venkata~Subrahmanyan Govindarajan, Benjamin~Van Durme, and Aaron~Steven White.
  2019.
\newblock Decomposing {{generalization}}: {{models}} of {{generic}},
  {{habitual}} and {{episodic statements}}.
\newblock \emph{Trans. Assoc. Comput. Linguistics}.

\bibitem[{Hershcovich et~al.(2018)Hershcovich, Abend, and
  Rappoport}]{MultitaskParsingSemantic_2018a}
Daniel Hershcovich, Omri Abend, and Ari Rappoport. 2018.
\newblock Multitask {{parsing across semantic representations}}.
\newblock In \emph{{{ACL}}}.

\bibitem[{Jiao et~al.(2023)Jiao, Wang, Huang, Wang, and
  Tu}]{ChatGPTGoodTranslator_2023}
Wenxiang Jiao, Wenxuan Wang, Jen-tse Huang, Xing Wang, and Zhaopeng Tu. 2023.
\newblock Is {{chatgpt a good translator}}? {{a preliminary study}}.
\newblock \emph{ArXiv}.

\bibitem[{Kipf and Welling(2017)}]{SemiSupervisedClassificationGraph_2017a}
Thomas~N. Kipf and Max Welling. 2017.
\newblock Semi-{{Supervised Classification}} with {{Graph Convolutional
  Networks}}.
\newblock In \emph{{{ICLR}} ({{Poster}})}.

\bibitem[{Li et~al.(2023)Li, Fang, Yang, Wang, Ye, Zhao, and
  Zhang}]{EvaluatingChatGPTInformation_2023}
Bo~Li, Gexiang Fang, Yang Yang, Quansen Wang, Wei Ye, Wen Zhao, and Shikun
  Zhang. 2023.
\newblock Evaluating {{chatgpt}}'s {{information extraction capabilities}}:
  {{an assessment}} of {{performance}}, {{explainability}}, {{calibration}},
  and {{faithfulness}}.
\newblock \emph{ArXiv}.

\bibitem[{Marcheggiani and Titov(2017)}]{EncodingSentencesGraph_2017a}
Diego Marcheggiani and Ivan Titov. 2017.
\newblock Encoding {{sentences}} with {{graph convolutional networks}} for
  {{semantic role labeling}}.
\newblock In \emph{{{EMNLP}}}.

\bibitem[{Moore and Lewis(2010)}]{IntelligentSelectionLanguage_2010a}
Robert~C. Moore and William~D. Lewis. 2010.
\newblock Intelligent {{selection}} of {{language model training data}}.
\newblock In \emph{{{ACL}}}.

\bibitem[{Oepen et~al.(2016)Oepen, Kuhlmann, Miyao, Zeman, Cinkov{\'a},
  Flickinger, Hajic, Ivanova, and
  Uresov{\'a}}]{ComparabilityLinguisticGraph_2016a}
Stephan Oepen, Marco Kuhlmann, Yusuke Miyao, Daniel Zeman, Silvie Cinkov{\'a},
  Dan Flickinger, Jan Hajic, Angelina Ivanova, and Zdenka Uresov{\'a}. 2016.
\newblock Towards {{comparability}} of {{linguistic graph banks}} for
  {{semantic parsing}}.
\newblock In \emph{{{LREC}}}.

\bibitem[{Oepen et~al.(2014)Oepen, Kuhlmann, Miyao, Zeman, Flickinger, Hajic,
  Ivanova, and Zhang}]{SemEval2014Task_2014a}
Stephan Oepen, Marco Kuhlmann, Yusuke Miyao, Daniel Zeman, Dan Flickinger, Jan
  Hajic, Angelina Ivanova, and Yi~Zhang. 2014.
\newblock {{Semeval}} 2014 {{task}} 8: {{broad-coverage semantic dependency
  parsing}}.
\newblock In \emph{{{SemEval}}@{{COLING}}}.

\bibitem[{Pennington et~al.(2014)Pennington, Socher, and
  Manning}]{GloveGlobalVectors_2014}
Jeffrey Pennington, Richard Socher, and Christopher~D. Manning. 2014.
\newblock Glove: {{global vectors}} for {{word representation}}.
\newblock In \emph{{{EMNLP}}}.

\bibitem[{Plank et~al.(2015)Plank, Alonso, Agic, Merkler, and
  S{\o}gaard}]{DependencyParsingMetrics_2015a}
Barbara Plank, H{\'e}ctor~Mart{\'i}nez Alonso, Zeljko Agic, Danijela Merkler,
  and Anders S{\o}gaard. 2015.
\newblock Do dependency parsing metrics correlate with human judgments?
\newblock In \emph{{{CoNLL}}}.

\bibitem[{Reisinger et~al.(2015)Reisinger, Rudinger, Ferraro, Harman, Rawlins,
  and Durme}]{SemanticProtoRoles_2015a}
Dee~Ann Reisinger, Rachel Rudinger, Francis Ferraro, Craig Harman, Kyle
  Rawlins, and Benjamin~Van Durme. 2015.
\newblock Semantic {{proto-roles}}.
\newblock \emph{Trans. Assoc. Comput. Linguistics}.

\bibitem[{Rudinger et~al.(2018)Rudinger, White, and
  Durme}]{NeuralModelsFactuality_2018}
Rachel Rudinger, Aaron~Steven White, and Benjamin~Van Durme. 2018.
\newblock Neural {{models}} of {{factuality}}.
\newblock In \emph{{{NAACL-HLT}}}.

\bibitem[{See et~al.(2017)See, Liu, and Manning}]{GetPointSummarization_2017}
Abigail See, Peter~J. Liu, and Christopher~D. Manning. 2017.
\newblock Get {{to the point}}: {{summarization}} with {{pointer-generator
  networks}}.
\newblock In \emph{{{ACL}}}.

\bibitem[{Silveira et~al.(2014)Silveira, Dozat, de~Marneffe, Bowman, Connor,
  Bauer, and Manning}]{GoldStandardDependency_2014a}
Natalia Silveira, Timothy Dozat, Marie-Catherine de~Marneffe, Samuel~R. Bowman,
  Miriam Connor, John Bauer, and Christopher~D. Manning. 2014.
\newblock A {{gold standard dependency corpus}} for {{english}}.
\newblock In \emph{{{LREC}}}.

\bibitem[{{Stengel-Eskin} et~al.(2021){Stengel-Eskin}, Murray, Zhang, White,
  and Durme}]{JointUniversalSyntactic_2021}
Elias {Stengel-Eskin}, Kenton~W. Murray, Sheng Zhang, Aaron~Steven White, and
  Benjamin~Van Durme. 2021.
\newblock Joint {{universal syntactic}} and {{semantic parsing}}.
\newblock \emph{Trans. Assoc. Comput. Linguistics}.

\bibitem[{{Stengel-Eskin} et~al.(2020){Stengel-Eskin}, White, Zhang, and
  Durme}]{UniversalDecompositionalSemantic_2020}
Elias {Stengel-Eskin}, Aaron~Steven White, Sheng Zhang, and Benjamin~Van Durme.
  2020.
\newblock Universal {{decompositional semantic parsing}}.
\newblock In \emph{{{ACL}}}.

\bibitem[{Sutskever et~al.(2014)Sutskever, Vinyals, and
  Le}]{SequenceSequenceLearning_2014}
Ilya Sutskever, Oriol Vinyals, and Quoc~V. Le. 2014.
\newblock Sequence to {{sequence learning}} with {{neural networks}}.
\newblock In \emph{{{NIPS}}}.

\bibitem[{Vashishtha et~al.(2019)Vashishtha, Durme, and
  White}]{FineGrainedTemporalRelation_2019a}
Siddharth Vashishtha, Benjamin~Van Durme, and Aaron~Steven White. 2019.
\newblock Fine-{{grained temporal relation extraction}}.
\newblock In \emph{{{ACL}}}.

\bibitem[{White et~al.(2016)White, Reisinger, Sakaguchi, Vieira, Zhang,
  Rudinger, Rawlins, and Durme}]{UniversalDecompositionalSemantics_2016}
Aaron~Steven White, Dee~Ann Reisinger, Keisuke Sakaguchi, Tim Vieira, Sheng
  Zhang, Rachel Rudinger, Kyle Rawlins, and Benjamin~Van Durme. 2016.
\newblock Universal {{decompositional semantics}} on {{universal
  dependencies}}.
\newblock In \emph{{{EMNLP}}}.

\bibitem[{White et~al.(2020)White, {Stengel-Eskin}, Vashishtha, Govindarajan,
  Reisinger, Vieira, Sakaguchi, Zhang, Ferraro, Rudinger, Rawlins, and
  Durme}]{UniversalDecompositionalSemantics_2020}
Aaron~Steven White, Elias {Stengel-Eskin}, Siddharth Vashishtha,
  Venkata~Subrahmanyan Govindarajan, Dee~Ann Reisinger, Tim Vieira, Keisuke
  Sakaguchi, Sheng Zhang, Francis Ferraro, Rachel Rudinger, Kyle Rawlins, and
  Benjamin~Van Durme. 2020.
\newblock The {{universal decompositional semantics dataset}} and {{decomp
  toolkit}}.
\newblock In \emph{{{LREC}}}.

\bibitem[{Wu et~al.(2023)Wu, Wang, Wan, Jiao, and
  Lyu}]{ChatGPTGrammarlyEvaluating_2023}
Haoran Wu, Wenxuan Wang, Yuxuan Wan, Wenxiang Jiao, and Michael Lyu. 2023.
\newblock {{Chatgpt}} or {{grammarly}}? {{evaluating chatgpt}} on {{grammatical
  error correction benchmark}}.
\newblock \emph{ArXiv}.

\bibitem[{Wu et~al.(2019)Wu, Wang, Xia, Qin, Lai, and
  Liu}]{ExploitingMonolingualData_2019a}
Lijun Wu, Yiren Wang, Yingce Xia, Tao Qin, Jianhuang Lai, and Tie-Yan Liu.
  2019.
\newblock Exploiting {{monolingual data}} at {{scale}} for {{neural machine
  translation}}.
\newblock In \emph{{{EMNLP}}/{{IJCNLP}}}.

\bibitem[{Zaremoodi et~al.(2018)Zaremoodi, Buntine, and
  Haffari}]{AdaptiveKnowledgeSharing_2018b}
Poorya Zaremoodi, Wray~L. Buntine, and Gholamreza Haffari. 2018.
\newblock Adaptive {{knowledge sharing}} in {{multi-task learning}}:
  {{improving low-resource neural machine translation}}.
\newblock In \emph{{{ACL}}}.

\bibitem[{Zhang et~al.(2020)Zhang, Zhang, Wang, Li, and
  Zhang}]{SyntaxAwareOpinionRole_2020a}
Bo~Zhang, Yue Zhang, Rui Wang, Zhenghua Li, and Min Zhang. 2020.
\newblock Syntax-{{aware opinion role labeling}} with {{dependency graph
  convolutional networks}}.
\newblock In \emph{{{ACL}}}.

\bibitem[{Zhang and Zong(2016)}]{ExploitingSourcesideMonolingual_2016}
Jiajun Zhang and Chengqing Zong. 2016.
\newblock Exploiting {{source-side monolingual data}} in {{neural machine
  translation}}.
\newblock In \emph{{{EMNLP}}}.

\bibitem[{Zhang et~al.(2019{\natexlab{a}})Zhang, Ma, Duh, and
  Durme}]{AMRParsingSequencetoGraph_2019}
Sheng Zhang, Xutai Ma, Kevin Duh, and Benjamin~Van Durme. 2019{\natexlab{a}}.
\newblock {{Amr parsing}} as {{sequence-to-graph transduction}}.
\newblock In \emph{{{ACL}}}.

\bibitem[{Zhang et~al.(2019{\natexlab{b}})Zhang, Ma, Duh, and
  Durme}]{BroadCoverageSemanticParsing_2019}
Sheng Zhang, Xutai Ma, Kevin Duh, and Benjamin~Van Durme. 2019{\natexlab{b}}.
\newblock Broad-{{coverage semantic parsing}} as {{transduction}}.
\newblock In \emph{{{EMNLP}}/{{IJCNLP}}}.

\bibitem[{Zhang et~al.(2017)Zhang, Rudinger, and
  Durme}]{EvaluationPredPattOpen_2017a}
Sheng Zhang, Rachel Rudinger, and Benjamin~Van Durme. 2017.
\newblock An {{evaluation}} of {{predpatt}} and {{open ie}} via {{stage}} 1
  {{semantic role labeling}}.
\newblock In \emph{{{IWCS}}}.

\bibitem[{Zhang et~al.(2018)Zhang, Qi, and
  Manning}]{GraphConvolutionPruned_2018a}
Yuhao Zhang, Peng Qi, and Christopher~D. Manning. 2018.
\newblock Graph {{convolution}} over {{pruned dependency trees improves
  relation extraction}}.
\newblock In \emph{{{EMNLP}}}.

\end{thebibliography}
\bibliographystyle{acl_natbib}

\end{document}